\documentclass[runningheads]{llncs}
\usepackage[section]{placeins}
\usepackage{graphicx}
\makeatletter 
  \newcommand\figcaption{\def\@captype{figure}\caption} 
  \newcommand\tabcaption{\def\@captype{table}\caption} 
\makeatother
\usepackage{algorithm}
\usepackage{multirow}
\usepackage{orcidlink}
\usepackage{listings}
\usepackage{xcolor}
\usepackage[normalem]{ulem}
\useunder{\uline}{\ul}{}
\usepackage{etoolbox}
\makeatletter
\AfterEndEnvironment{algorithm}{\let\@algcomment\relax}
\AtEndEnvironment{algorithm}{\kern2pt\hrule\relax\vskip3pt\@algcomment}
\let\@algcomment\relax
\newcommand\algcomment[1]{\def\@algcomment{\footnotesize#1}}
\renewcommand\fs@ruled{\def\@fs@cfont{\bfseries}\let\@fs@capt\floatc@ruled
  \def\@fs@pre{\hrule height.8pt depth0pt \kern2pt}%
  \def\@fs@post{}%
  \def\@fs@mid{\kern2pt\hrule\kern2pt}%
  \let\@fs@iftopcapt\iftrue}
\makeatother
\usepackage{booktabs}
\usepackage{bbding}
\usepackage{tikz}
\usepackage{comment}
\usepackage{amsmath,amssymb} 
\usepackage{color}
\usepackage{gensymb}
\usepackage[accsupp]{axessibility}  

\begin{document}
\pagestyle{headings}
\mainmatter
\def\ECCVSubNumber{2463}  

\title{GaitEdge: Beyond Plain End-to-end Gait Recognition for Better Practicality} 

\titlerunning{GaitEdge}
%
\author{Junhao Liang\inst{1}$^\star$
\orcidlink{0000-0001-5612-7631} 
\and Chao Fan\inst{1}\thanks{Equal contributions.}
\orcidlink{0000-0002-3605-2705} 
\and Saihui Hou\inst{2,4}
\orcidlink{0000-0003-4689-2860} 
\and Chuanfu Shen\inst{3,1}
\orcidlink{0000-0001-8782-5950} 
\and Yongzhen Huang\inst{2,4}
\orcidlink{0000-0003-4389-9805} 
\and Shiqi Yu
\inst{1(}\Envelope\inst{)}
\orcidlink{0000-0002-5213-5877}
}
\authorrunning{J. Liang et al.}
%
\institute{Southern University of Science and Technology \\
\and
School of Artificial Intelligence, Beijing Normal University \\
\and
The University of Hong Kong \\
\and
WATRIX.AI
}
\maketitle

\begin{abstract}
Gait is one of the most promising biometrics to identify individuals at a long distance. 
Although most previous methods have focused on recognizing the silhouettes, several end-to-end methods that extract gait features directly from RGB images perform better. 
However, we demonstrate that these end-to-end methods may inevitably suffer from the gait-irrelevant noises, \textit{i.e.}, low-level texture and colorful information.
Experimentally, we design  the \textbf{cross-domain} evaluation to support this view.
In this work, we propose a novel end-to-end framework named \textbf{GaitEdge} which can effectively block gait-irrelevant information and release end-to-end training potential. 
Specifically, GaitEdge synthesizes the output of the pedestrian segmentation network and then feeds it to the subsequent recognition network, where the synthetic silhouettes consist of trainable edges of bodies and fixed interiors to limit the information that the recognition network receives. 
Besides, \textbf{GaitAlign} for aligning silhouettes is embedded into the GaitEdge without losing differentiability. 
Experimental results on CASIA-B and our newly built TTG-200 indicate that GaitEdge significantly outperforms the previous methods and provides a more practical end-to-end paradigm. All the source code are available at \url{https://github.com/ShiqiYu/OpenGait}.
\keywords{Gait Recognition; End-to-end; Gait Alignment; Cross-domain}
\end{abstract}

\section{Introduction}

In recent years, human identification by walking pattern, \textit{i.e.}, gait, has become a hot research topic.
Compared with other biometrics, \textit{e.g.}, face, fingerprint, and iris, 
human gait can be easily captured at a long distance without the cooperation of subjects, 
which means gait can be promising for crimes investigation and suspects tracing under real-world uncontrolled conditions.
It is noticed that most of the studies treat gait recognition as a two-step approach, 
including extracting the intermediate modality, \textit{e.g.}, silhouette mask or skeleton keypoints, from RGB images
and putting them into the downstream gait recognition network.
However, some researches~\cite{chao2019gaitset,fan2020gaitpart,hou2020gait,lin2021gait} indicate that those multi-step pipelines usually give rise to the weakness in  efficiency and effectiveness; increasing works tend to infer the final results directly in end-to-end~\cite{zhang2019gait,song2019gaitnet,li2020end}.

\begin{figure}[tb]
\centering
\includegraphics[height=5cm]{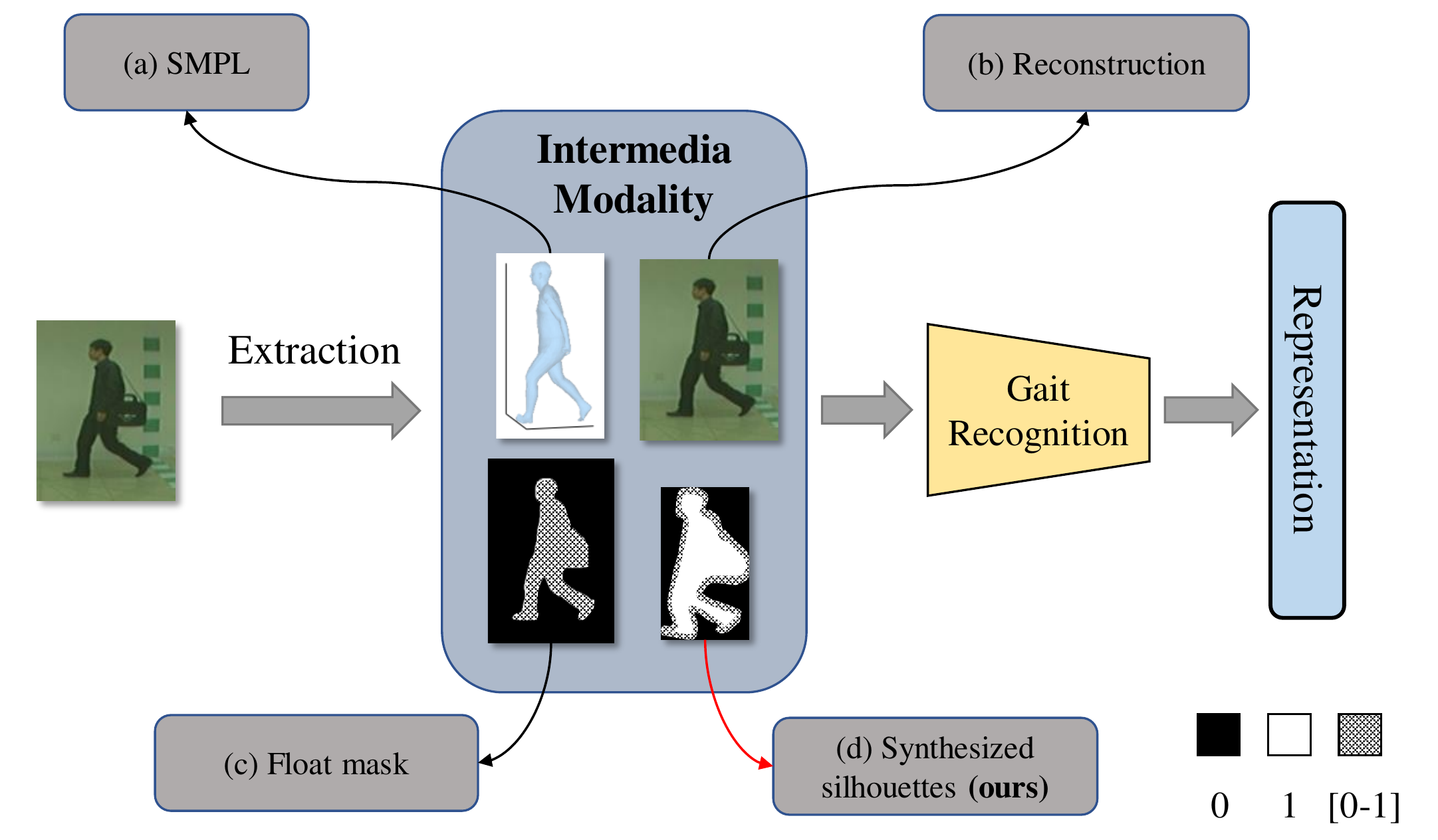}
\caption{Three typical end-to-end approaches: (a) model-based end-to-end~\cite{li2020end,li2021end}, (b) Zhang's GaitNet~\cite{zhang2019gait}, (c) Song's GaitNet~\cite{song2019gaitnet} and (d) our \textbf{GaitEdge}. \textit{Shaded areas} for the float-point numbers ranging from 0 to 1 }
\label{fig:intro}
\end{figure}

To the best of our knowledge, 
there are three typical end-to-end gait recognition methods in the recent literature.
As illustrated in Figure \ref{fig:intro} (a),
Li \textit{et al.}~\cite{li2020end,li2021end} utilize a fashion human mesh recovery model~\cite{kanazawa2018end} to reconstruct the three-dimensional human body and train the recognition network by taking the parameters of the skinned multi-person linear (SMPL)~\cite{loper2015smpl} model as inputs.
Another typical approach proposed by Zhang \textit{et al.}~\cite{zhang2019gait} introduces an autoencoder framework to disentangle the motion-relevant gait patterns and motion-irrelevant appearance features explicitly from the sequential RGB images, as shown in Figure \ref{fig:intro} (b).
In addition, Song \textit{et al.} propose GaitNet~\cite{song2019gaitnet}, which integrates two tasks, \textit{i.e.},  pedestrian segmentation and gait recognition, as illustrated in Figure \ref{fig:intro} (c). 
It extracts gait features straightly from the intermediate float mask instead of the classical binary silhouettes.

Although achieving higher performance than those two-step methods,
we argue that these end-to-end methods cannot ensure that the learned characteristics for human identification only consist of walking patterns. 
Since  the intermediate modalities of previous end-to-end frameworks,  
\textit{e.g.}, the SMPL reconstruction in~\cite{li2020end,li2021end}, 
posture features disentanglement in~\cite{zhang2019gait}, 
and pedestrian segmentation supervision in~\cite{song2019gaitnet}, are float-coding, they may introduce some background and texture information.
Furthermore, while the previous methods all attempt to exclude gait-irrelevant features, they lack convincing  experiments to validate.
To alleviate these issues, 
we notice that gait features
are generally more robust for the covariants of camera viewpoints, carrying, and clothing than other gait-irrelevant noises, \textit{i.e.}, texture and color,
which implies that if these uncorrelated features dominate in the extracted gait representations, the recognition performance will drop a lot when the model is directly exploited in unseen domains (new dataset)~\cite{huang2018eanet}.
Hence, in this paper, we introduce the \textbf{cross-domain} evaluation to expose the side effects of RGB information.
More importantly, we propose a concise yet compelling end-to-end framework named \textbf{GaitEdge} to deal with this challenging evaluation.
As shown in Figure \ref{fig:intro} (d), the intermediate modality of GaitEdge is a novel synthetic silhouette, 
while its edge is composed of the trainable float mask, and other regions are classical binary silhouettes.
Two intuitive phenomena inspire this design:
First, the RGB-informed noises are mainly distributed in the non-edge regions, \textit{e.g.}, the human body and background.
Therefore, treating these regions as binary silhouettes can effectively prevent the leakage of gait-irrelevant noises. 
Second, the edge region plays a vital role in describing the shape of the human body.
Hence, making the only edge region trainable is enough to liberate the potential of the end-to-end training strategy.
In addition, 
we observe that the size-normalized alignment~\cite{iwama2012isir} is necessary for the silhouette pre-processing to keep the body in aspect ratio. Unfortunately, this operation used to be offline and thus non-differentiable, which means it can not be directly applied to align the synthetic silhouette.
To solve this problem, inspired by the RoIAlign~\cite{he2017mask}, 
we propose \textbf{GaitAlign} module to complete the framework of GaitEdge, which can be regarded as a differentiable version of the alignment method proposed by~\cite{iwama2012isir}.

In summary, we make the following three major contributions: 
(1) We point out concerns about gait-irrelevant noises being mixed into the final gait representations and introduce cross-domain testing to verify the leakage of RGB-informed noises. Besides, due to the lack of a gait dataset providing RGB videos, we collect the \textit{Ten Thousand Gaits} (TTG-200), whose size is approximately equal to the popular CASIA-B~\cite{yu2006framework}. 
(2) We propose the GaitEdge, a concise yet compelling end-to-end gait recognition framework. Experiments on both CASIA-B and TTG-200 demonstrate that GaitEdge reaches a new state-of-the-art performance, and we declare that GaitEdge can effectively prevent irrelevant RGB-informed noises.
(3) We propose a module named GaitAlign for the silhouette-based end-to-end gait recognition, and it can be considered a differentiable version of size-normalization~\cite{iwama2012isir}.

\section{Related Work}
\subsection{Gait Recognition}

As a kind of biometrics, gait is defined by early research~\cite{winter1991biomechanics} as the walking pattern that a given person will perform in a fairly repeatable and characteristic way. 
On the other hand, another similar task, \textit{i.e.}, person re-identification~\cite{zheng2016person}, aims to find a person presented in one camera in another place by another camera. 
Despite the similarity, they are still fundamentally different: the first task focuses on walking patterns, while the second task uses clothing primarily for identification. 
Therefore, it is worth emphasizing that we can not let the gait recognition network acquire information other than gait patterns, such as RGB-informed texture and color.

At present, the mainstream visual-based gait recognition methods can be roughly divided into model-based  and appearance-based.  The former model-based approaches~\cite{liao2020model,an2020performance,li2020end,teepe2021gaitgraph} usually extract the underlying structure of the human body first, 
\textit{e.g.}, 2D or 3D skeleton key-points,
and then model the human walking patterns.
In general, such methods can better mitigate the effects of clothing and more accurately describe the body's posture. 
Nonetheless, all of them are difficult to model the human body structure under the practical surveillance scene due to the low quality of the video.

More and more appearance-based gait recognition methods~\cite{wu2016comprehensive,zhang2019cross,chao2019gaitset,zhang2020learning,fan2020gaitpart,hou2020gait,lin2021gait} are currently leaving model-based methods behind. 
Recently, GaitSet~\cite{chao2019gaitset} takes a sequence of silhouettes as input and makes great progress.
Subsequently, Fan \textit{et al.}~\cite{fan2020gaitpart} propose a focal convolutional layer to learn the  part-level feature and  utilize Micro-motion Capture Module to model short-range temporal patterns. 
Besides, Lin \textit{et al.}~\cite{lin2021gait} propose a 3D CNN-based Global and Local Feature Extractor module to extract discriminative global and local representations from frames, which outperforms the other methods remarkably. 



\subsection{End-to-end Learning}
End-to-end learning refers to integrating several separate gradient-based deep learning modules in a differentiable manner. This training paradigm has a natural advantage in that the system optimizes components for overall performance rather than optimizing human-selected intermediates~\cite{bojarski2016end}. 

Recently, some excellent research has benefited from the end-to-end learning paradigm. Amodei \textit{et al.}~\cite{amodei2016deep} replace entire pipelines of hand-engineered components with neural networks to overcome the diverse variety of speech by end-to-end learning. Another notable work~\cite{bojarski2016end} is Nvidia's end-to-end training for autonomous driving systems. It only gives the system the human steering angle as the training signal. Still, the system can automatically learn the internal representation of the necessary processing steps, such as detecting lane lines. 

As the end-to-end philosophy becomes increasingly popular, several studies~\cite{zhang2019gait,li2020end,song2019gaitnet} have applied it to gait recognition. 
Firstly, Zhang \textit{et al.}~\cite{zhang2019gait} propose an autoencoder to disentangle the appearance and gait information without the explicit appearance and gait label for supervision. 
Secondly, Li \textit{et al.}~\cite{li2020end,li2021end} use the newly developed 3D human mesh model~\cite{loper2015smpl} as an intermediate modality and make the silhouettes generated by the neural 3D mesh renderer~\cite{kato2018neural} consistent with the silhouette segmented from RGB images. 
Because the 3D mesh model provides more helpful information than silhouettes, this approach achieves state-of-the-art results. 
However, using a 3D mesh model requires a higher resolution of an input RGB image, which is not feasible in the real surveillance scenario. 
Different from the previous two, Song \textit{et al.}~\cite{song2019gaitnet} propose another type of end-to-end gait recognition framework. 
It is formed by directly connecting the pedestrian segmentation and gait recognition networks, which is supervised by a joint loss, \textit{i.e.}, segmentation loss and recognition loss. 
This approach looks relatively more applicable, but it is also likely to result in the gait-irrelevant noises leaking into the recognition network due to the absence of explicit restrictions. 
Under this consideration, Our GaitEdge mainly poses and addresses two pivotal problems: cross-domain evaluation and silhouette misalignment.

\section{Cross Domain Problem}
\begin{figure}[tb]
\centering
\includegraphics[height=6.5cm]{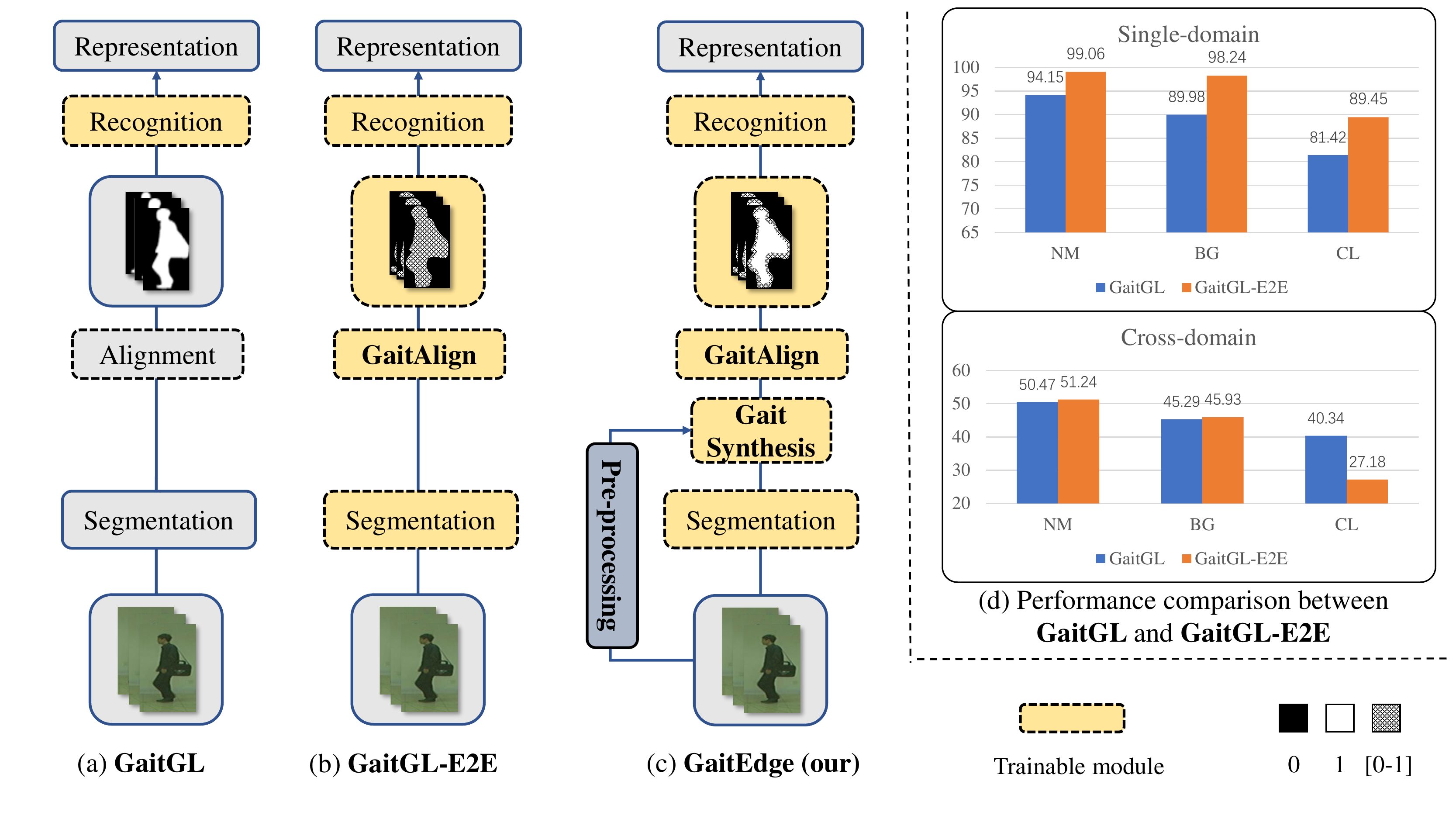}
\caption{(a), (b) and (c) are three different frameworks. (d) The rank-1 accuracy (\%) on CASIA-B* excluding identical-view cases. \textit{NM} for normal walking, \textit{BG} for walking with bags, and \textit{CL} for walking with cloth change}
\label{fig:cross-domain}
\end{figure}

From the previous perspective, we argue that although the existing end-to-end approaches~\cite{zhang2019gait,song2019gaitnet,li2020end,li2021end} greatly improve the accuracy, it is natural to suspect that the introduction of RGB information is the cause of the improvement. To verify our conjecture, we introduce two gait recognition paradigms and compare them experimentally.

Firstly, one of the best-performing two-step gait recognition methods, \textit{i.e.}, \textbf{GaitGL}~\cite{lin2021gait}, is adopted as a baseline. In addition, a simple and straightforward end-to-end model named \textbf{GaitGL-E2E} that provides a fair comparison is introduced. As shown in Figure \ref{fig:cross-domain} (a) and (b), both methods use the same modules except that GaitGL-E2E replaces binary mask with float-coding silhouettes through a trainable segmentation network, \textit{i.e.}, U-Net~\cite{ronneberger2015u}. 
Experimentally, we define the \textbf{single-domain} evaluation as training and testing on CASIA-B*\footnote{We reprocess CASIA-B and denote the newly processed one as CASIA-B*.}~\cite{yu2006framework}. Correspondingly, the \textbf{cross-domain} evaluation is defined as training on another dataset, \textit{i.e.}, TTG-200, but testing the trained model on CASIA-B*. More implementation details will be elaborated in Section \ref{sec:exp}. 

As shown in the single-domain part of Figure \ref{fig:cross-domain} (d), GaitGL-E2E easily outperforms GaitGL because it has more trainable parameters, and more information is contained in the float-point mask than the binary mask. However, it is inevitable to doubt that float-point numbers flowing into the recognition network bring in texture and color from RGB images, which makes the recognition network learn gait-irrelevant information and leads to the degradation of cross-domain performance.
On the other hand, the cross-domain part of Figure \ref{fig:cross-domain} (d) shows that GaitGL-E2E does not achieve the same advantages as it does in single-domain and is even much lower than GaitGL (GaitGL: 40.34\%, GaitGL-E2E: 27.18\%) in the most challenging case, \textit{i.e.}, CL (walking with cloth change). This phenomenon indicates it is easier for end-to-end models to learn easily identifiable coarse-grained RGB information rather than fine-grained imperceptible gait patterns.

The above two experiments demonstrate that GaitGL-E2E does absorb RGB noises so that it is no longer reliable for gait recognition with practical cross-domain requirements. Therefore, we propose a novel framework GaitEdge composed of our carefully designed Gait Synthesis module and differentiable GaitAlign module, as shown in Figure \ref{fig:cross-domain} (c). The most significant difference between GaitEdge and GaitGL-E2E is that we control the transmission of RGB information through manual silhouettes synthesis. 

\section{Our Framework}

\begin{figure}[tb]
\centering
\includegraphics[height=4.5cm]{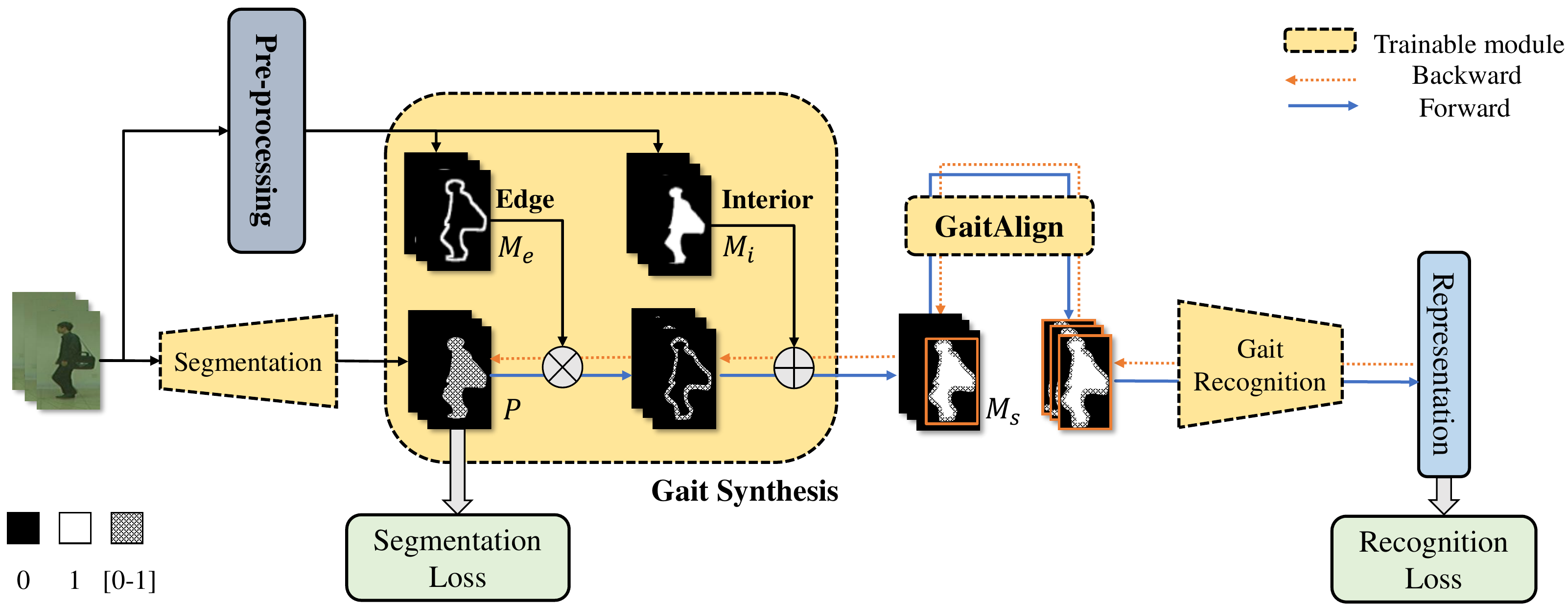}
\caption{Illustration of GaitEdge, $\oplus$ for \textit{add element-wise}, $\otimes$ for \textit{multiply element-wise}, and \textit{shaded areas} for float-point numbers ranging from 0 to 1. More details about \textbf{Pre-processing} module can be found in Figure \ref{fig:preproc}.}
\label{fig:pipeline}
\end{figure}

\label{sec:method}

\subsection{Gait Synthesis}
We generally believe that the edge (the contour of the silhouette) contains the most discriminative information in silhouette images~\cite{yu2007reducing}. The interior of a silhouette can be regarded as low-frequency content with less information, whereas the information will be too compact  to train the recognition network if we get rid of the interior. Therefore, the designed module, named \textbf{Gait Synthesis}, focuses on combining trainable edges with fixed interiors through mask operation. It only trains the edge part of the silhouette image, and the region other than edges are extracted from the frozen segmentation network. As shown in Figure \ref{fig:pipeline}, to clarify how our framework works, we use yellow for the trainable module and illustrate the flow of gradient transfer, in which the dotted orange line represents the backward propagation, and the solid blue line represents the forward propagation. The masks of edge and interior are denoted as $M_{e}$ and $M_{i}$. The output probability of the segmentation network is denoted as $P$. Then, the output of Gait Synthesis denoted $M_s$ can be obtained by several element-wise operations:
\begin{equation}
\label{eq:syn}
M_s=M_e \times P + M_i
\end{equation}
As shown in Equation \ref{eq:syn}, we explicitly multiply $P_s$ by $M_{e}$ and then add it to $M_{i}$, which blocks most information, including the gait-relevant and gait-irrelevant. However, we can still fine-tune the edges of the silhouettes, making it automatically optimized for recognition.

\begin{figure}[b]
\centering
\includegraphics[height=2cm]{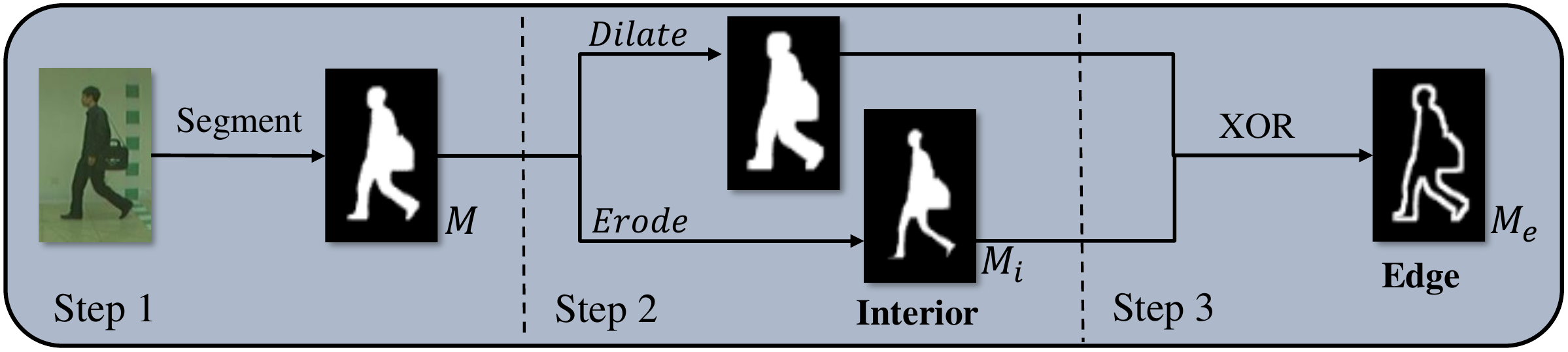}
\caption{The \textbf{Pre-processing} module in GaitEdge}
\label{fig:preproc}
\end{figure}

\subsubsection{Pre-processing.} We design an untrainable pre-processing operation to get $M_{e}$ and $M_{i}$, shown in Figure \ref{fig:preproc}. Specifically, we divide it to three steps. First, we segment the input RGB image with the trained segmentation model to obtain the silhouette $M$. Then, in the second step, we use the classic morphological algorithms to get the dilation and erosion silhouettes($M_i$) with a $3\times 3$ flat structuring element. Finally, we get $M_e$ by element-wise exclusive or $\veebar$. Formally:
\begin{equation}
\begin{aligned}
M_i &=erode(M) \\
M_e &=M_i \veebar dilate(M) 
\end{aligned}
\end{equation}


Overall, \textit{Gait Synthesis} takes the most intuitive approach by limiting the adjustable region to retain the most valuable silhouette features while eliminate most of low-level RGB-informed noises. It is worth mentioning that, \textit{Gait Synthesis} can be detachably integrated into previous silhouette-based end-to-end methods due to the simplicity of the design.

\subsection{Gait Alignment Module}
\label{sec:align}
Alignment is very crucial for all silhouette-based gait recognition methods. Since the size-normalization of the silhouette was used for the first time on the OU-ISIR Gait Database~\cite{iwama2012isir}, almost all silhouette-based methods pre-process the silhouette input via size-normalization, which removes the noise and benefits the recognition. However, the previous end-to-end approach, \textit{i.e.}, GaitNet~\cite{song2019gaitnet}, feeds the segmented silhouette into the recognition network directly,  which hardly handles the situation mentioned above. Therefore, we propose a differentiable Gait Alignment Module called \textit{GaitAlign} to make the body be the center of the image and fill the entire image vertically.

\begin{algorithm}[tb]
\caption{Pseudocode of GaitAlign in a PyTorch-like style.}
\label{alg:align}
\algcomment{\fontsize{8pt}{0em}\selectfont \texttt{bbox}: Get the four regularly locations of bounding box keeping aspect ratio. We hide this tedious engineering trick, and the source code will be released. 

\texttt{roi\_align}: Crop and resize the interested region without the loss of spatial alignment.
}
\definecolor{codeblue}{rgb}{0.25,0.5,0.5}
\lstset{
  backgroundcolor=\color{white},
  basicstyle=\fontsize{8pt}{8pt}\ttfamily\selectfont,
  columns=fullflexible,
  breaklines=true,
  captionpos=b,
  commentstyle=\fontsize{8pt}{8pt}\color{codeblue},
  keywordstyle=\fontsize{8pt}{8pt},
}

\begin{lstlisting}[language=python]
# s_in  : silhouettes from segmentation output, (nx1xhxw)
# size  : the target size, (H,W)
# r : aspect ratio of human body, (n)
# s_out : aligned silhouettes, (nx1xHxW)

# pad along the x axis so as not to exceed the boundary
s_in = ZeroPad2d((w // 2, w // 2), 0, 0)) # (nx1xhx2w)
binary_mask = round(s_in) # binary silhouette

# compute the coordinates and restore the aspect ratio r
left, top, right, bottom = bbox(binary_mask, r, size)

# get the new silhouettes by differentiable roi_align
s_out = roi_align(s_in, (left, top, right, bottom), size)
\end{lstlisting}
\end{algorithm}

We first review the size-normalization ~\cite{iwama2012isir} procedure because GaitAlign can be regarded as a differentiable version. In size-normalization, by figuring out the top, bottom, and horizontal center of the body, we can scale the body to the target height in aspect ratio and then pad the x-axis with zeros to reach the target width. In our case, pseudo-code in Algorithm \ref{alg:align} depicts the procedure of GaitAlign.
We first need to pad the left and right sides with half the width of zeros, which ensures that crop operation will not exceed the boundary. According to the aspect ratio and the target size, then we compute the exact values of four regularly sampled locations. Finally, RoIAlign ~\cite{he2017mask} is applied to the locations given by the previous step. As a result, we get the standard-size, image-filled silhouettes, and its aspect ratio remains the same (refer to the output of GaitAlign in Figure \ref{fig:pipeline}). Another noteworthy point is that the GaitAlign module is still \textbf{differentiable}, making our end-to-end training feasible.



%





\section{Experiment}
\label{sec:exp}

\subsection{Settings}

\subsubsection{Datasets.} 
There are a few available datasets for gait recognition, \textit{e.g.}, CASIA-B~\cite{yu2006framework}, OUMVLP~\cite{takemura2018multi},  Outdoor-Gait~\cite{song2019gaitnet}, FVG~\cite{zhang2019gait}, GREW~\cite{zhu2021gait}, and so on.
However, not all of them are useful for the end-to-end based gait recognition methods.
For example, the proposed work cannot apply the two worldwide most enormous gait datasets, OUMVLP~\cite{takemura2018multi} and GREW~\cite{zhu2021gait}, because neither provides RGB videos.
In short, our ideal gait dataset owns several vital attributes: RGB videos available, rich camera viewpoints, and multiple walking conditions. 

\begin{figure}[tb]
\centering
\includegraphics[height=2cm]{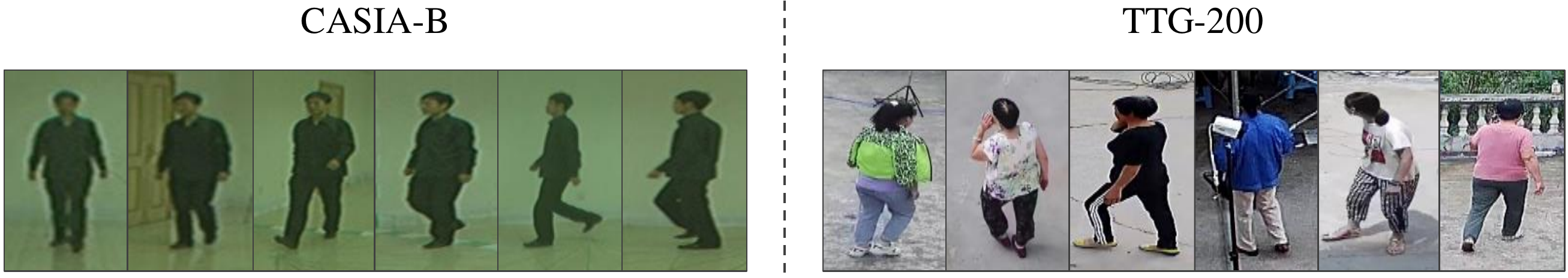}
\caption{Examples of CASIA-B and TTG-200. The left (CASIA-B) consists of six views of one sequence. The right (TTG-200) consists of six subjects with different views}
\label{fig:dataset}
\end{figure}
CASIA-B~\cite{yu2006framework} seems to be a good choice. 
Nevertheless, there still needs another similar dataset to fit our cross-domain settings.
Consequently, we collect a private dataset named \textit{Ten Thousand Gaits 200 (TTG-200)} and show its statistics in Table \ref{tab:dataset}.

\begin{table}[b]
\centering
\caption{The statistics of existing gait datasets and our collected TTG-200}
\label{tab:dataset}
\resizebox{\linewidth}{!}{
\begin{tabular}{c|c|c|c|c|c}
\hline
Dataset      & Subjects & Environment & Format     & Variations                                                                                                   & \#Sequnces \\ \hline
CASIA-B      & 124      & Indoor      & RGB        & 11 views, carrying, clothing                                                                                 & 13,636      \\ \hline
OUMVLP       & 10,307    & Indoor      & Silhouette & 14 views                                                                                                     & 267,388     \\ \hline
FVG          & 226      & Outdoor     & RGB        & \begin{tabular}[c]{@{}c@{}}3 frontal views, walking speed, \\ carrying, clothing, background\end{tabular} & 2,856       \\ \hline
Outdoor-Gait & 138      & Outdoor     & RGB        & carrying, clothing                                                                                           & 4,964       \\ \hline
GREW         & 26,345    & Outdoor     & Silhouette & multiple camera                                                                                              & 128,671     \\ \hline
\textbf{TTG-200 (our)}      & 200      & Outdoor     & RGB        & \begin{tabular}[c]{@{}c@{}}12 views, carrying, clothing, \\ talking on the phone, \\ background\end{tabular} & 14,198      \\ \hline
\end{tabular}
}
\end{table}

\paragraph{CASIA-B.} There are 124 subjects walking indoor in CASIA-B. 
It is probably the most popular dataset that consists of 11 views ([0\degree-180\degree]) and three walking conditions, \textit{i.e.}, normal walking (NM\#01-06), walking with bags (BG\#01-02), and walking with cloth change (CL\#01-02). 
We strictly follow previous studies, which group the first 74 subjects into the training set and others into the test set. 
Furthermore, for the test stage, the first 4 sequences (NM\#01-04) are regarded as the gallery set, while the left 6 sequences are grouped into 3 probe subsets, \textit{i.e.}, NM\#05-06, BG\#01-02, CL\#01-02.
Besides, since the silhouettes of CASIA-B were obtained by the outdated background subtraction, there exists much noise caused by the background and clothes of subjects.
Hence, we re-annotate the silhouettes of CASIA-B and denote it as \textit{CASIA-B*}. All our experiments are conducted on this newly annotated one. 

\paragraph{TTG-200.} 
This dataset contains 200 subjects walking in the wild, and each subject is required to walk under 6 various conditions, \textit{i.e.}, carrying, clothing, taking on the phone, and so on.
For each walking process, the subject will be captured by 12 cameras located around the different viewpoints (unlabelled), which means each subject ideally owns  $6 \times 12 = 72$ gait sequences.
In the following experiments, 
we take the first 120 subjects for training and the last 80 subjects for the test. 
In addition, the first sequence (\#1) is regarded as gallery set, and the left 5 sequences (\#2-6) are regarded as probe set. 

As shown in Figure \ref{fig:dataset}, compared with CASIA-B, TTG-200 has three main differences: (1) The backgrounds of TTG-200 are more complex and diverse (collected in multiple different outdoor scenes); (2) The data of TTG-200 are mostly aerial view, while data of CASIA-B are mostly horizontal view; (3) TTG-200 has better image quality. 
Therefore, we can treat these two datasets as different domains.

\subsubsection{Implementation Details}
\label{sec:impl}
\paragraph{Data Pre-processing.}
We first employ ByteTrack~\cite{zhang2021bytetrack} to detect and track pedestrians from the raw RGB videos for both CASIA-B~\cite{yu2006framework} and TTG-200,
and then conduct the human segmentation and silhouette alignment~\cite{iwama2012isir} to extract the gait sequences. 
The obtained silhouettes are resized to $64\times44$ and can be taken as the input of these two-stage gait recognition methods or be the ground-truth for the pedestrian segmentation network in these end-to-end based approaches.

\paragraph{Pedestrian Segmentation.}
We use the popular U-Net~\cite{ronneberger2015u} as our segmentation network that is supervised by Binary Cross-Entropy~\cite{jadon2020survey} loss $L_{seg}$. We set the input size as $128\times128\times3$ and the channels of U-Net as \{3, 16, 32, 64, 128, 64, 32, 16, 1\} and train it via SGD~\cite{ruder2016overview} (batch size=960, momentum=0.9, initial learning rate=0.1, weight decay=$5\times10^{-4}$). For each dataset, we train the network with learning rate scaled to 1/10 two times for every 10000 iterations until convergence.

\paragraph{Gait Recognition.}
We use the latest GaitGL~\cite{lin2021gait} as our recognition network and strictly follow the original paper's settings.

\paragraph{Joint Training Details.}
In this step, the training data sampler and batch size are similar to the gait recognition network.
We jointly fine-tune the segmentation and recognition networks with the joint loss $L_{joint}=\lambda L_{seg}+L_{rec}
$,
where $L_{rec}$ denotes the loss of recognition network.
The $\lambda$ represents the loss weight of segmentation network and is set to 10.
Besides, to make the joint training process converge faster, 
we use the trained segmentation and recognition networks parameters to initialize the end-to-end model, 
and accordingly, their initial learning rate is set to $10^{-5}$ and $10^{-4}$, respectively.
Moreover, we fix the first half of the segmentation network, \textit{i.e.}, U-Net, to keep the segmentation result in human shape.
We jointly train the end-to-end network for a total of 20,000 iterations and reduce the learning rate by 1/10 at the 10,000th iteration.

\subsection{Performance Comparison}
To demonstrate the reliable cross-domain capability of GaitEdge, we conduct the single-domain and cross-domain evaluations on CASIA-B* and TTG-200, as shown in Table \ref{tab:overall}.


The diagonal of Table \ref{tab:overall} shows the single-domain performance comparisons, 
where these methods are trained and evaluated in the identical dataset.
On the opposite, the anti-diagonal shows the cross-domain performance comparisons, 
where these methods are trained and evaluated in the different datasets.

\begin{table}[tb]
\centering
\caption{The rank-1 accuracy (\%) on CASIA-B* and TTG-200. The identical-view cases in CASIA-B* are excluded. The \textbf{bold} and \textbf{(bold)} numbers for the two highest accuracies of single-domain and that of cross-domain, respectively}
\label{tab:overall}
\resizebox{0.7\linewidth}{!}{
\begin{tabular}{c|c|c|c|c|c|c|c|c} 
\hline
\multirow{3}{*}{Training Set} & \multicolumn{2}{c|}{\multirow{3}{*}{Method}}     & \multicolumn{6}{c}{Test Set}                                                                                                         \\ 
\cline{4-9}
                              & \multicolumn{2}{c|}{}                            & \multicolumn{4}{c|}{CASIA-B*}                                                             &                      & TTG-200           \\ 
\cline{4-7}\cline{9-9}
                              & \multicolumn{2}{c|}{}                            & NM                   & BG                   & CL                   & Mean              &                      & -                 \\ 
\cline{1-7}\cline{9-9}
\multirow{5}{*}{CASIA-B*}     & \multirow{3}{*}{Two-step} & GaitSet~\cite{chao2019gaitset}              & 92.30                & 86.10                & 73.36                & 83.92                &                      & 40.26             \\
                              &                           & GaitPart~\cite{fan2020gaitpart}            & 93.14                & 85.99                & 75.05                & 84.72                &                      & 42.23             \\
                              &                           & GaitGL~\cite{lin2021gait}              & 94.15                & 89.98                & 81.42                & 88.52                &                      & \textbf{(48.74)}  \\ 
\cline{2-7}\cline{9-9}
                              & \multirow{2}{*}{End2end}  & GaitGL-E2E           & 99.06                & 98.24                & 89.45                & \textbf{95.58}       &                      & 37.18             \\
                              &                           & \textbf{GaitEdge}    & 97.94                & 96.06                & 86.36                & \textbf{93.45}       &                      & \textbf{(49.12)}  \\ 
\cline{1-7}\cline{9-9}
\multicolumn{1}{c}{}          & \multicolumn{1}{c}{}      & \multicolumn{1}{c}{} & \multicolumn{1}{c}{} & \multicolumn{1}{c}{} & \multicolumn{1}{c}{} & \multicolumn{1}{c}{} & \multicolumn{1}{c}{} &                   \\ 
\cline{1-7}\cline{9-9}
\multirow{5}{*}{TTG-200}      & \multirow{3}{*}{Two-step} & GaitSet~\cite{chao2019gaitset}             & 41.32                & 35.15                & 21.59                & 32.69                &                      & 77.62             \\
                              &                           & GaitPart~\cite{fan2020gaitpart}            & 45.21                & 38.75                & 25.92                & 36.62                &                      & 80.24             \\
                              &                           & GaitGL~\cite{lin2021gait}              & 50.47                & 45.29                & 40.34                & \textbf{(45.37)}     &                      & 80.46             \\ 
\cline{2-7}\cline{9-9}
                              & \multirow{2}{*}{End2end}  & GaitGL-E2E           & 51.24                & 45.93                & 27.18                & 41.45                &                      & \textbf{90.37}    \\
                              &                           & \textbf{GaitEdge}    & 54.76                & 49.85                & 38.16                & \textbf{(47.59)}     &                      & \textbf{88.66}    \\
\hline
\end{tabular}
}
\end{table}
\subsubsection{Single-domain Evaluation.} 
From the diagonal results in Table \ref{tab:overall}, we observe that the performance of traditional two-step gait recognition methods is far inferior to that of two end-to-end ones. For example, GaitGL-E2E exceeds GaitSet~\cite{chao2019gaitset} by 11.66\% for CASIA-B* and 12.75\% for TTG-200, respectively.
On the other hand, the accuracy of our proposed GaitEdge is slightly lower than that of GaitGL-E2E, \textit{i.e.}, -2.13\% for CASIA-B* and -1.71\% for TTG-200.
However, we argue that GaitGL-E2E owns the higher risk of overfitting in the gait-irrelevant noises since it directly takes the float mask generated by the segmentation network as the input of the recognition network.
Hence, we further conduct the cross-domain evaluation to to support this notion experimentally. 

\subsubsection{Cross-domain Evaluation.} 
If some irrelevant noises dominate the gait representations used for 
human identification, 
\textit{i.e.}, texture and color, 
the recognition accuracy would drop dramatically in the case of cross-domain settings since the extracted features impotently represent the relatively robust gait patterns.
The anti-diagonal results in Table \ref{tab:overall} show that all these methods have significant  performance degradation compared to single-domain due to the significant difference between CASIA-B* and TTG-200. 
We notice that although GaitGL-E2E has the highest accuracy in single-domain, it achieves the poorest performance for crossing the domain from CASIA-B* to TTG-200. 
In contrast, our GaitEdge reaches the best performance than any other posted method in cross-domain evaluations, although it is about 2\%  lower than GaitGL-E2E in single domain.

Hence, this cross-domain evaluation not only indicates the robustness of GaitEdge is far superior to that of GaitGL-E2E 
but also claims the GaitEdge is a practical and advanced framework for the end-to-end gait recognition task.

\subsubsection{Comparison with other end-to-end methods.}
\begin{table}[tb]
\centering
\caption{The rank-1 accuracy (\%) on CASIA-B* across different views excluding the identical-view cases. For evaluation, the first 4 sequences (NM\#01-04) are regarded as the gallery set, while the left 6 sequences are grouped into 3 probe subsets, \textit{i.e.}, NM\#05-06, BG\#01-02, CL\#01-02. The original paper of Song GaitNet~\cite{song2019gaitnet} does not mention the results of BG and CL}
\label{tab:end2end}
\resizebox{0.9\linewidth}{!}{%
\begin{tabular}{c|c|c|c|c|c|c|c|c|c|c|c|c|c} 
\hline
\multirow{2}{*}{Probe} & \multirow{2}{*}{Method} & \multicolumn{11}{c|}{Probe View}                                                                                                                                              & \multirow{2}{*}{Mean}  \\ 
\cline{3-13}
                       &                         & 0°            & 18°           & 36°           & 54°           & 72°           & 90°           & 108°          & 126°          & 144°          & 162°          & 180°          &                        \\ 
\hline
\multirow{5}{*}{NM}    & Song GaitNet~\cite{song2019gaitnet}            & 75.6          & 91.3          & 91.2          & 92.9          & 92.5          & 91.0          & 91.8          & 93.8          & 92.9          & 94.1          & 81.9          & 89.9                   \\
                       & Zhang GaitNet~\cite{zhang2019gait}           & 93.1          & 92.6          & 90.8          & 92.4          & 87.6          & 95.1          & 94.2          & 95.8          & 92.6          & 90.4          & 90.2          & 92.3                   \\
                       & ModelGait~\cite{li2020end}               & 96.9          & 97.1          & 98.5          & 98.4          & \textbf{97.7} & 98.2          & 97.6          & 97.6          & 98.0          & 98.4          & \textbf{98.6} & 97.9                   \\
                       & MvModelGait~\cite{li2021end}             & \textbf{97.5} & 97.6          & 98.6          & \textbf{98.8} & \textbf{97.7} & \textbf{98.9} & \textbf{98.9} & 97.3          & 97.6          & 97.8          & 97.9          & \textbf{98.1}          \\
                       & \textbf{GaitEdge}       & 97.2          & \textbf{99.1} & \textbf{99.2} & 98.3          & 97.3          & 95.5          & 97.1          & \textbf{99.4} & \textbf{99.3} & \textbf{98.5} & 96.4          & 97.9                   \\ 
\hline
\multirow{4}{*}{BG}    & Zhang GaitNet~\cite{zhang2019gait}           & 88.8          & 88.7          & 88.7          & 94.3          & 85.4          & \textbf{92.7} & 91.1          & 92.6          & 84.9          & 84.4          & 86.7          & 88.9                   \\
                       & ModelGait~\cite{li2020end}                & 94.8          & 92.9          & 93.8          & 94.5          & 93.1          & 92.6          & \textbf{94.0} & 94.5          & 89.7          & 93.6          & 90.4          & 93.1                   \\
                       & MvModelGait~\cite{li2021end}             & 93.9          & 92.5          & 92.9          & 94.1          & 93.4          & 93.4          & 95.0          & 94.7          & 92.9          & 93.1          & 92.1          & 93.4                   \\
                       & \textbf{GaitEdge}       & \textbf{95.3} & \textbf{97.4} & \textbf{98.4} & \textbf{97.6} & \textbf{94.3} & 90.6          & 93.1          & \textbf{97.8} & \textbf{99.1} & \textbf{98.0} & \textbf{95.0} & \textbf{96.1}          \\ 
\hline
\multirow{4}{*}{CL}    & Zhang GaitNet~\cite{zhang2019gait}           & 50.1          & 60.7          & 72.4          & 72.1          & 74.6          & 78.4          & 70.3          & 68.2          & 53.5          & 44.1          & 40.8          & 62.3                   \\
                       & ModelGait~\cite{li2020end}                & 78.2          & 81.0          & 82.1          & 82.8          & 80.3          & 76.9          & 75.5          & 77.4          & 72.3          & 73.5          & 74.2          & 77.6                   \\
                       & MvModelGait~\cite{li2021end}             & 77.0          & 80.0          & 83.5          & 86.1          & 84.5          & \textbf{84.9} & 80.6          & 80.4          & 77.4          & 76.6          & 76.9          & 80.7                   \\
                       & \textbf{GaitEdge}       & \textbf{84.3} & \textbf{92.8} & \textbf{94.3} & \textbf{92.2} & \textbf{84.6} & 83.0          & \textbf{83.0} & \textbf{87.5} & \textbf{87.4} & \textbf{85.9} & \textbf{75.0} & \textbf{86.4}          \\
\hline
\end{tabular}
}
\end{table}
Last but not least, the proposed GaitEdge is compared to three previous end-to-end gait recognition methods across different views on CASIA-B*. Table \ref{tab:end2end}
shows that GaitEdge reaches  almost the highest accuracy on various walking conditions, especially for CL (+5.7\% than MvModelGait), which reveals that GaitEdge is remarkably robust to color and texture (cloth change).
\begin{table}[b]
\centering
\caption{The ablation study for the size of structuring element. The larger size for the larger edge region. The \textbf{bold} and \textbf{(bold)} numbers for the highest accuracy of single-domain and that of cross-domain, respectively}
\label{tab:struct}
\resizebox{0.7\linewidth}{!}{
\begin{tabular}{c|c|c|c|c|c|c|l|c} 
\hline
\multirow{3}{*}{Training Set} & \multirow{3}{*}{Method}            & \multirow{3}{*}{\begin{tabular}[c]{@{}c@{}}Structuring\\ Element\end{tabular}} & \multicolumn{6}{c}{Test Set}                                                                                                              \\ 
\cline{4-9}
                              &                                    &                                                                                & \multicolumn{4}{c|}{CASIA-B*}                                                             & \multicolumn{1}{c|}{} & TTG-200               \\ 
\cline{4-7}\cline{9-9}
                              &                                    &                                                                                & NM                   & BG                   & CL                   & Mean              &                       & -                     \\ 
\cline{1-7}\cline{9-9}
\multirow{5}{*}{CASIA-B*}     & \multirow{4}{*}{\textbf{GaitEdge}} & $3\times 3$                                                                    & 97.94                & 96.06                & 86.36                & 93.45                &                       & 49.12                 \\
                              &                                    & $5\times 5$                                                                    & 98.88                & 97.36                & 88.24                & 94.83                &                       & \textbf{(50.98)}      \\
                              &                                    & $7\times 7$                                                                    & 98.97                & 97.90                & 88.36                & 95.08                &                       & 49.15                 \\
                              &                                    & $9\times 9$                                                                    & 99.02                & 98.19                & 89.05                & 95.42                &                       & 44.47                 \\ 
\cline{2-7}\cline{9-9}
                              & GaitGL-E2E                         & -                                                                              & 99.06                & 98.24                & 89.45                & \textbf{95.58}       &                       & 37.18                 \\ 
\cline{1-7}\cline{9-9}
\multicolumn{1}{l}{}          & \multicolumn{1}{l}{}               & \multicolumn{1}{l}{}                                                           & \multicolumn{1}{l}{} & \multicolumn{1}{l}{} & \multicolumn{1}{l}{} & \multicolumn{1}{l}{} & \multicolumn{1}{l}{}  & \multicolumn{1}{l}{}  \\ 
\cline{1-7}\cline{9-9}
\multirow{5}{*}{TTG-200}      & \multirow{4}{*}{\textbf{GaitEdge}} & $3\times 3$                                                                    & 54.76                & 49.85                & 38.16                & \textbf{(47.59)}     &                       & 88.66                 \\
                              &                                    & $5\times 5$                                                                    & 49.21                & 45.22                & 33.71                & 42.71                &                       & 89.62                 \\
                              &                                    & $7\times 7$                                                                    & 50.26                & 44.20                & 32.43                & 42.29                &                       & 90.00                 \\
                              &                                    & $9\times 9$                                                                    & 48.84                & 41.72                & 27.66                & 39.41                &                       & \textbf{90.39}        \\ 
\cline{2-7}\cline{9-9}
                              & GaitGL-E2E                         & -                                                                              & 51.24                & 45.93                & 27.18                & 41.45                &                       & 90.37                 \\
\hline
\end{tabular}
}
\end{table}
\subsection{Ablation Study}
\subsubsection{Impact of Edge.}
Table \ref{tab:struct} shows the impact of body edge size. We extract the edges by several sizes of structuring elements—the larger the structuring element, the larger the edge area. According to the results shown in Table \ref{tab:struct}, as the size of the structuring element increases, the performance of single-domain accordingly increases, 
but the performance of cross-domain almost decreases at the same time. 
This result claims that the area of the float mask occupying the intermediate synthetic silhouette is negatively associated with the cross-domain performance for GaitEdge.
Therefore, we can argue that the reason why GaitGL-E2E fails in cross-domain evaluation is that it is equivalent to GaitEdge in the case of the infinite structural element. Furthermore, those non-edge regions of silhouette, \textit{i.e.}, human body and background, are unsuitable in float-coding for the end-to-end gait recognition framework.

\begin{figure}[tb]
\centering
\includegraphics[height=4.5cm]{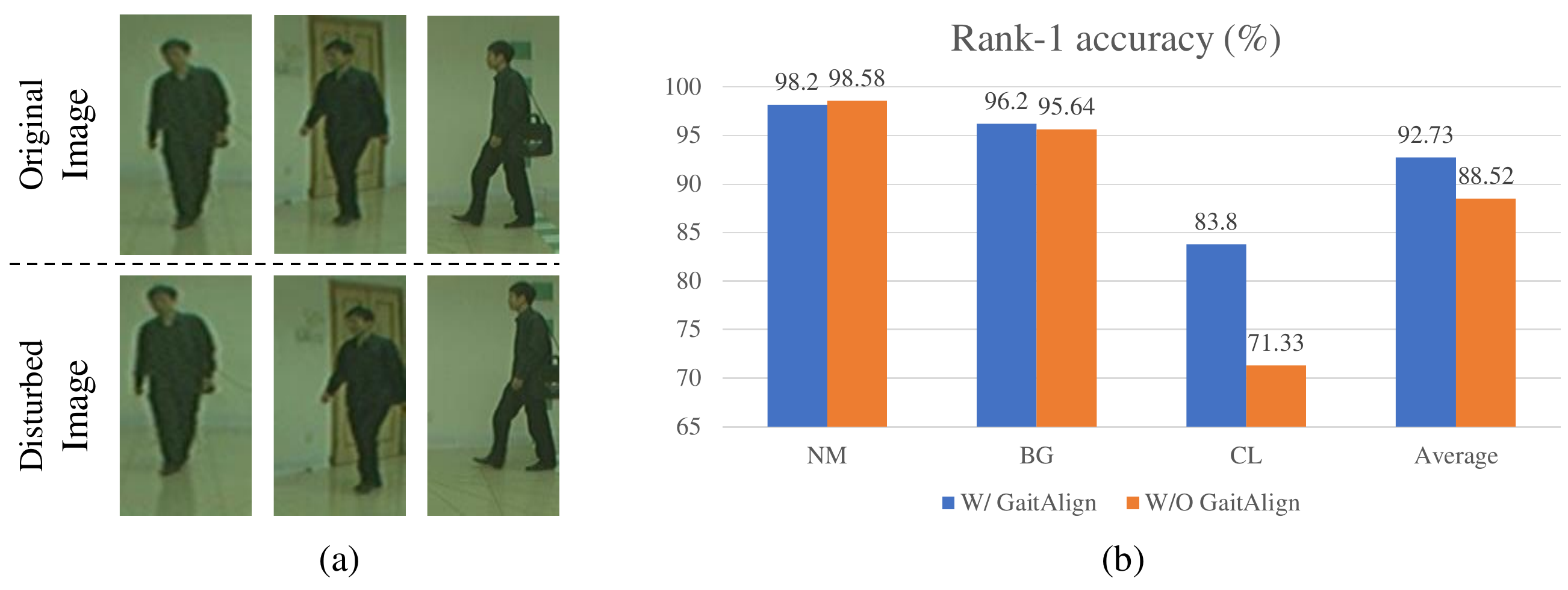}
\caption{(a) The original images (top) \textit{vs.} the disturbed images (bottom). We make random pixel offset to the disturbed images, including vertical and horizontal direction. (b) The ablation study for the \textit{GaitAlign} module. The results are reported on CASIA-B* after disturbance}
\label{fig:align}
\end{figure}
\subsubsection{Impact of GaitAlign.}
Notably, we observe that the result of pedestrian detection (upstream task) in natural scenes, is often much worse than that of the controlled environment, \textit{i.e.,} CASIA-B* and TTG-200. 
In order to simulate this complex situation, we first apply object detection on the videos of CASIA-B* and then perform random pixel offset along with vertical and horizontal coordinates with a probability of 0.5. 
As shown in Figure \ref{fig:align} (a), the bottom images are disturbed, aiming to simulate the  natural situation. 
Figure \ref{fig:align} (b) shows that alignment improves the average accuracy significantly. In addition, we also notice that the accuracy of normal walking (NM) drops a little, \textit{i.e.}, -0.38\%. However, we believe this is because the accuracy of NM is approaching the upper limit.
 

\subsection{Visualization}
\begin{figure}[tb]
\centering
\includegraphics[height=3.5cm]{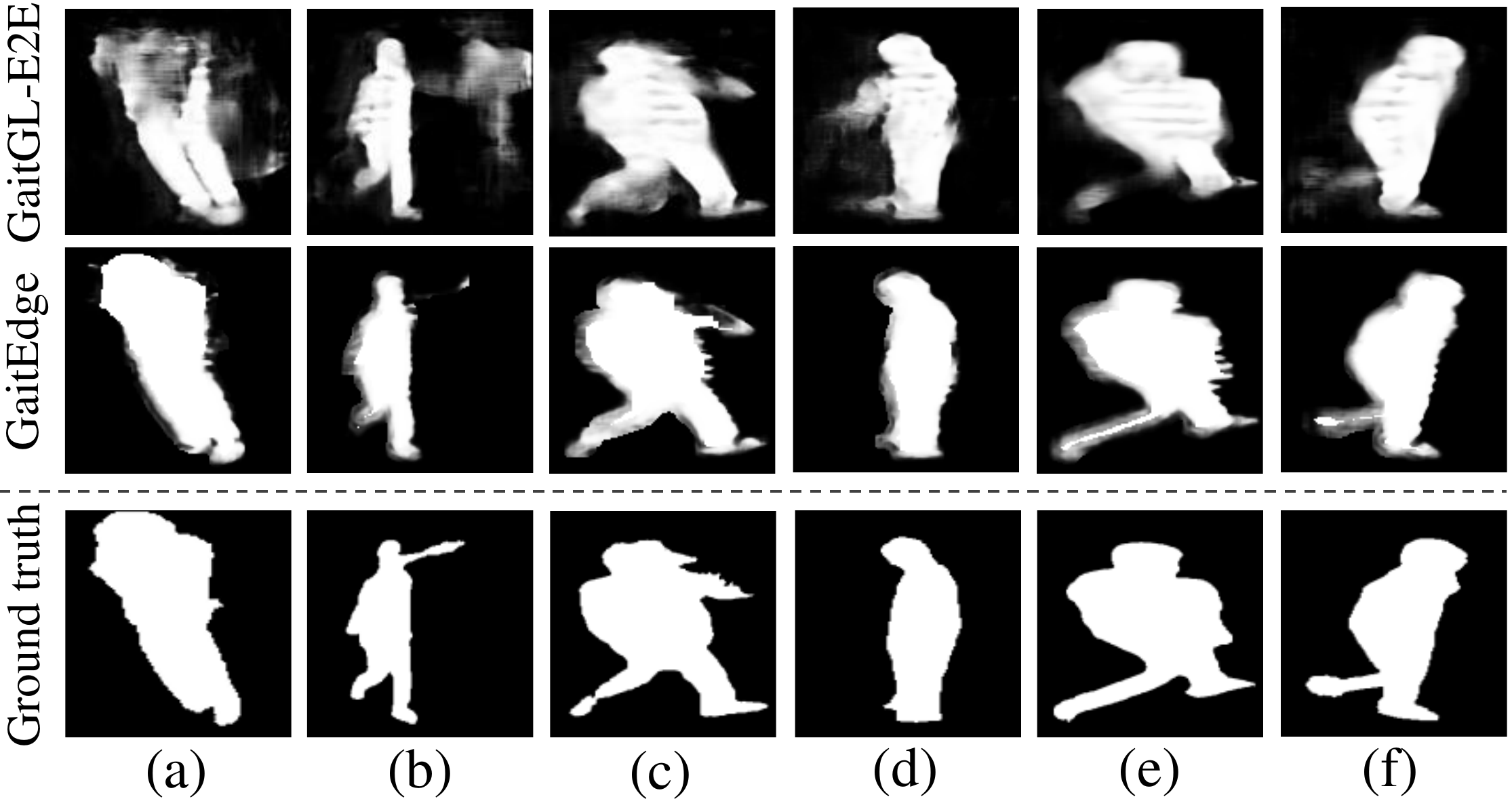}
\caption{The comparison between the intermediate results and ground truth, \textit{i.e.}, the first two rows \textit{v.s.} the third row}
\label{fig:viz}
\end{figure}

To better understand the performance degradation of GaitGL-E2E and the effectiveness of GaitEdge, we illustrate the intermediate results generated by GaitGL-E2E and GaitEdge respectively as well as the ground truth corresponding to the same frame, as shown in Figure \ref{fig:viz}.
Specifically, for GaitGL-E2E, the intermediate results in (a), (b), (c) and (d) capture more background and texture information, and some body parts are eliminated such as legs in (e) and (f).
While for GaitEdge, the intermediate results are much more stable and reasonable making it more robust.


\section{Conclusion}

This paper presents a novel end-to-end gait recognition framework termed GaitEdge that can solve the performance degradation in cross-domain situation. Specifically, we design a Gait Synthesis module to mask the fixed body with tunable edges obtained by morphological operation. Besides, a differentiable alignment module named GaitAlign is proposed to solve the body position jitter caused by the upstream pedestrian detection task. We also conduct extensive and comprehensive experiments on two datasets, including CASIA-B* and our newly built TTG-200. Experimental results show that GaitEdge significantly outperforms the previous methods, indicating that GaitEdge is a more practical end-to-end paradigm that can effectively block RGB noise. Moreover, this work exposes the cross-domain problem neglected by previous studies, which provides a new perspective for future research.

~\\
\textbf{Acknowledgments} We would like to thank the helpful discussion with Dr. Chunshui Cao and Dr. Xu Liu.
This work was supported in part by the National Natural Science Foundation of China under Grant 61976144, 
in part by the Stable Support Plan Program of Shenzhen Natural Science Fund under Grant 20200925155017002, 
in part by the National Key Research and Development Program of China under Grant 2020AAA0140002, 
and in part by the Shenzhen Technology Plan Program (Grant No. KQTD20170331093217368).

\par\vfill\par

\clearpage
%
%
\bibliographystyle{splncs04}
\bibliography{egbib}
\end{document}